\documentclass{article} 
\usepackage{iclr2026_conference,times}

\usepackage{amsmath,amsfonts,bm}

\def\eqref#1{equation~\ref{#1}}

\def\1{\bm{1}}

\DeclareMathAlphabet{\mathsfit}{\encodingdefault}{\sfdefault}{m}{sl}
\SetMathAlphabet{\mathsfit}{bold}{\encodingdefault}{\sfdefault}{bx}{n}
\newcommand{\tens}[1]{\bm{\mathsfit{#1}}}

\def\tH{{\tens{H}}}

\def\tW{{\tens{W}}}

\usepackage{hyperref}
\usepackage{url}
\usepackage{booktabs}
\usepackage{multirow}
\usepackage{array}
\usepackage{listings}
\usepackage{xcolor}
\usepackage{graphicx}
\usepackage[most]{tcolorbox}
\usepackage{pgf-pie}
\usepackage{tikz}
\usetikzlibrary{arrows.meta}
\usetikzlibrary{shapes.geometric, positioning}
\usepackage{subcaption}
\usepackage{float}
\usepackage{adjustbox}
\usepackage{graphicx}
\usepackage{stfloats} 
\usepackage{makecell}
\usepackage[section]{placeins} 

\usepackage{pgfplots}
\usepgfplotslibrary{statistics}
\usepgfplotslibrary{polar}
\pgfplotsset{compat=1.18}
\usepackage{caption}
\newcommand{\nobrk}[1]{\mbox{#1}}

\usepackage{tikz}
\usepackage{amsmath}
\usepackage{xcolor}
\usepackage{graphicx}
\usetikzlibrary{calc,positioning,fit,arrows.meta}

\title{Spatial Reasoning is Not a Free Lunch: \\
A Controlled Study on LLaVA}

\author{%
\normalfont\mdseries
Nahid Alam$^{1,\dagger}$,
Leema Krishna Murali$^{1,5,\dagger}$,
Siddhant Bharadwaj$^{2}$,
Patrick Liu$^{3,\dagger}$, \\
Timothy Chung$^{4,1}$,
Drishti Sharma$^{1}$,
Akshata A.$^{1,\dagger}$,
Kranthi Kiran$^{1,6}$, \\
Wesley Tam$^{6,\dagger}$,
Bala Krishna S Vegesna$^{7}$\\
\\
$^1$Cohere Labs Community,
$^2$Indian Institute of Science, Bangalore,
$^3$UIUC,\\
$^4$Imperial College London,
$^5$Eisai Inc.,
$^6$EleutherAI,
$^7$Georgia Institute of Technology\\[2pt]
\\
$\dagger$ Work done as part of the EleutherAI SOAR Program\\
\texttt{\small nahid.m.alam@gmail.com}
}

\iclrfinalcopy 
\begin{document}

\maketitle

\begin{abstract}
Vision-language models (VLMs) have advanced rapidly, yet they still struggle with basic spatial reasoning. Despite strong performance on general benchmarks, modern VLMs remain brittle at understanding 2D spatial relationships such as relative position, layout, and counting. We argue that this failure is not merely a data problem, but is closely tied to dominant design choices in current VLM pipelines: reliance on CLIP-style image encoders and the flattening of images into 1D token sequences with 1D positional encoding. We present a controlled diagnostic study within the LLaVA framework to isolate how these choices affect spatial grounding. We evaluate frontier models and LLaVA variants on a suite of spatial benchmarks, comparing CLIP-based encoders against alternatives trained with denser or generative objectives, as well as variants augmented with 2D positional encoding. Our results show consistent spatial performance gaps across models, and indicate that encoder objectives and positional structure shape spatial behavior, but do not fully resolve it.
\end{abstract}

\section{Background and Motivation}
\label{sec:background}

Modern vision-language models (VLMs) almost universally rely on large pre-trained image encoders such as CLIP and SigLIP \citep{dosovitskiy2021imageworth16x16words, radford2021learning, sun2023evaclipimprovedtrainingtechniques, oquab2024dinov2learningrobustvisual, zhai2023sigmoid, tschannen2025siglip2multilingualvisionlanguage}. These encoders are trained primarily to align global image representations with text and are then integrated into systems such as Flamingo, LLaVA, BLIP-2, KOSMOS, Florence-2, and Molmo \citep{alayrac2022flamingo, liu2023llava, liu2023improvedllava, li2023blip, peng2023kosmos, kosmos-g, xiao2024florence, deitke2024molmo}. While this paradigm has driven progress on captioning and VQA, it does not explicitly optimize for structured spatial representations. Prior work shows that CLIP-style encoders emphasize semantic alignment while underperforming on fine-grained and spatially grounded tasks \citep{tong2024eyeswideshutexploring, anis2025limitationsvisionlanguagemodelsunderstanding}. Recent encoders introduce denser or generative objectives \citep{oquab2024dinov2learningrobustvisual, maninis2025tipstextimagepretrainingspatial, tschannen2025siglip2multilingualvisionlanguage, fini2024multimodalautoregressivepretraininglarge}, but their impact on spatial grounding in VLMs remains underexplored.

Beyond the encoder, multimodal alignment introduces a second structural bottleneck. Most VLMs flatten images into 1D token sequences before applying 1D rotary positional encoding \citep{su2023roformerenhancedtransformerrotary}, collapsing 2D structure during fusion. Recent analyses argue that this design undermines spatial reasoning even when strong visual features are available \citep{zhang2025scalingbeyondadvancingspatial, zhang2025mllmsstrugglespatialunderstanding}. While Qwen2-VL introduces multimodal rotary embeddings that preserve height and width information \citep{wang2024qwen2}, systematic evidence isolating the role of positional structure remains limited.

At the evaluation level, spatial reasoning is rarely treated as a first-class capability. Foundational VLMs are primarily reported on general benchmarks, while spatial understanding is often omitted \citep{alayrac2022flamingo, liu2023llava, li2023blip, xiao2024florence}. Meanwhile, several benchmarks and datasets now explicitly target spatial reasoning, including MMVP, CV-Bench, GQA, VSR, TopViewRS, TallyQA, and CountBenchQA \citep{tong2024eyeswideshutexploring, tong2024cambrian1, hudson2019gqa, liu2023vsr, li2024topviewrs, acharya2018tallyqa, paliGemma2024}, as well as training-driven efforts such as RoboSpatial, SpatialVLM, MM-Spatial, and SpatialRGPT \citep{song2025robospatialteachingspatialunderstanding, chen2024spatialvlmendowingvisionlanguagemodels, daxberger2025mmspatialexploring3dspatial, cheng2024spatialrgptgroundedspatialreasoning}. However, these efforts often conflate data, scale, and architecture, making it difficult to isolate which design factors shape spatial behavior.

In this work, we focus on static 2D spatial reasoning and present a controlled diagnostic study within the LLaVA framework. We evaluate frontier VLMs and systematically vary two under-examined design dimensions: the image encoder objective and the positional structure used during multimodal alignment, introducing 2D rotary positional encoding. This setting allows us to isolate how these factors influence spatial grounding and to assess the extent to which they account for observed spatial failures.

\section{Experimental Setup}
\label{sec:experiment}

\subsection{Methods}
All experiments are conducted within the LLaVA \citep{liu2023llava} framework. We construct controlled LLaVA-1.5 (7B) variants by swapping the image encoder while holding the language backbone and training pipeline fixed. Specifically, we compare CLIP, SigLIP, SigLIP2, and AIMv2, and evaluate each with and without 2D rotary positional encoding (2D-RoPE).

Unlike standard 1D RoPE, 2D-RoPE encodes both horizontal and vertical patch indices and is applied to query and key projections during multimodal attention. This preserves explicit 2D structure during image–text fusion and allows us to isolate the effect of positional structure on spatial grounding.

\newcommand{\TrapW}{0.5333cm}   
\newcommand{\TrapTW}{0.3667cm}
\newcommand{\TrapH}{1.05cm}
\newcommand{\CanvasW}{13.8cm}

\tikzset{
  pics/trap/.style n args={2}{
      code={
          \coordinate (#2_bc) at (0,0);
          \coordinate (#2_bl) at ($ (#2_bc) + (-0.5*\TrapW, 0) $);
          \coordinate (#2_br) at ($ (#2_bc) + ( 0.5*\TrapW, 0) $);
          \coordinate (#2_tl) at ($ (#2_bc) + (-0.5*\TrapTW, \TrapH) $);
          \coordinate (#2_tr) at ($ (#2_bc) + ( 0.5*\TrapTW, \TrapH) $);
          \coordinate (#2_tc) at ($ (#2_tl)!0.5!(#2_tr) $);
          \draw[fill=#1, draw=black, line width=0.3pt]
          (#2_bl) -- (#2_br) -- (#2_tr) -- (#2_tl) -- cycle;
        }
    }
}
\begin{figure}[h]
  \centering
  \begin{tikzpicture}[scale=0.7, every node/.style={scale=0.7}, >=Stealth, ]

    \newcommand{\TopY}{3.0cm}
    \newcommand{\BotY}{0.30cm}
    \newcommand{\TopStep}{1.0cm}
    \newcommand{\BotStep}{1.0cm}

    \newcommand{\LLMH}{1cm}
    \pgfmathsetlengthmacro{\LLMbaseY}{(\BotY+\TrapH+\TopY-\LLMH)/2}
    \coordinate (llmSW) at (0,\LLMbaseY);
    \coordinate (llmNE) at (\CanvasW,{\LLMbaseY+\LLMH});
    \draw[rounded corners=6pt, fill=green!12, draw=green!60!black, line width=0.6pt]
    (llmSW) rectangle (llmNE);

    \node[anchor=west, align=left]
    at ($(llmSW)+(0.45cm,0.45*\LLMH)$) {\large \textbf{Large Language Model} $f_{\phi}$};

    \begin{scope}[shift={(0.4cm,-1.7cm)}]
      \newcommand{\ImgW}{2cm}
      \newcommand{\ImgH}{2cm}
      \newcommand{\Gap}{0.06cm}

      \pgfmathsetlengthmacro{\tW}{\ImgW/3}
      \pgfmathsetlengthmacro{\tH}{\ImgH/3}

      \foreach \i in {0,1,2}{
          \foreach \j in {0,1,2}{
              \pgfmathsetlengthmacro{\xshift}{\i*(\tW+\Gap)}
              \pgfmathsetlengthmacro{\yshift}{\j*(\tH+\Gap)}
              \pgfmathsetlengthmacro{\xoff}{-\i*\tW}
              \pgfmathsetlengthmacro{\yoff}{-\j*\tH}

              \begin{scope}[shift={(\xshift,\yshift)}]
                \clip (0,0) rectangle (\tW,\tH);
                \node[anchor=south west, inner sep=0pt]
                at (\xoff,\yoff)
                {\includegraphics[width=\ImgW,height=\ImgH]{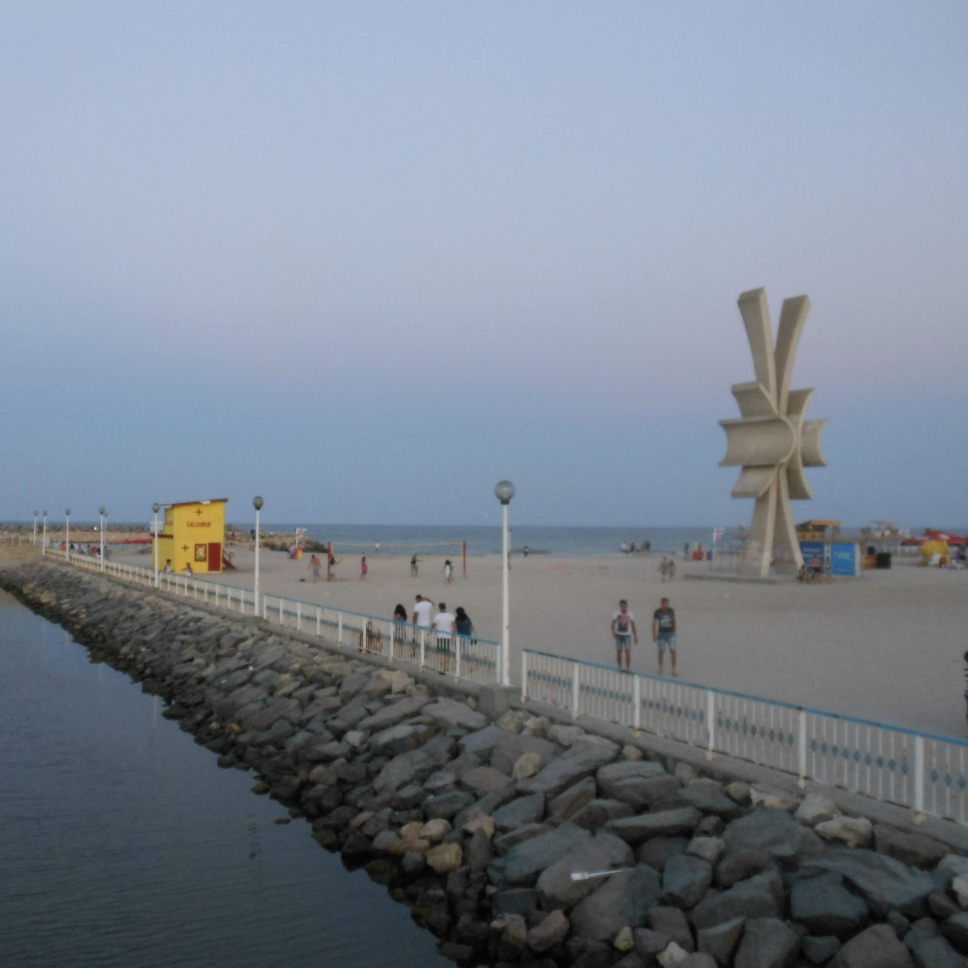}};
                \draw[black, line width=0.4pt, rounded corners=1pt] (0,0) rectangle (\tW,\tH);
              \end{scope}
            }
        }

      \pgfmathsetlengthmacro{\PWw}{3*\tW + 2*\Gap}
      \pgfmathsetlengthmacro{\PWh}{3*\tH + 2*\Gap}
      \coordinate (pwSW) at (0,0);
      \coordinate (pwSE) at (\PWw,0);
      \coordinate (pwNW) at (0,\PWh);
      \coordinate (pwNE) at (\PWw,\PWh);
      \coordinate (pwBC) at ($ (pwSW)!0.5!(pwSE) $);
      \coordinate (pwTC) at ($ (pwNW)!0.5!(pwNE) $);

      \node[below=10pt] (imgcap) at (pwBC) {Image $\mathbf{X}_v$};
      \draw[->, thin] (imgcap.north) -- (pwBC);

      \node[draw=none, fill=none, inner sep=4pt]
      (imgenc) at ($(pwTC)+(0,6mm)$) {};

      \node[draw=cyan!60!black, rounded corners=3pt, fill=cyan!25,
        inner sep=4pt, anchor=west, align=left]
      at ($(llmSW)+(0cm,-0.5cm)$)  
      (imgencvis) {\small Image Encoder (CLIP/SigLIP/SigLIP2/AIMv2)};
      \draw[->, thin] (pwTC) -- (imgenc.south);

      \node[draw, rounded corners=3pt, fill=yellow!35, inner sep=4pt, below right=5.2mm and 1.4cm of imgenc]
      (projw) {\small Projection $\mathbf{W}$};
      \draw[->, thin] (projw) ++(0,1cm) -- (projw);

    \end{scope}

    \coordinate (rEdge) at (\CanvasW,0cm);
    \newcommand{\RightBC}[2]{($(rEdge)+(-0.5*\TrapW - #2, #1)$)}

    \pic at \RightBC{\TopY}{0cm}            {trap={green!60!black!25}{A}};
    \pic at \RightBC{\TopY}{\TopStep}       {trap={green!60!black!25}{B}};
    \pic at \RightBC{\TopY}{2*\TopStep}     {trap={green!60!black!25}{C}};
    \node[align=center] at ($ (A_tc)!0.5!(C_tc) + (-3.5,-0.65cm) $)
    {$\mathbf{X}_a$: \textit{Language Response}};

    \newcommand{\ShiftLeft}{6.0cm}
    \pic at \RightBC{\BotY}{\ShiftLeft + 0cm}        {trap={yellow!35}{E}};
    \pic at \RightBC{\BotY}{\ShiftLeft + \BotStep}   {trap={yellow!35}{F}};
    \pic at \RightBC{\BotY}{\ShiftLeft + 2*\BotStep} {trap={yellow!35}{G}};
    \node[below=4pt] (HvLabel) at ($ (E_bc)!0.5!(G_bc) $) {$\mathbf{H}_v$};

    \newcommand{\ShiftRight}{2cm}
    \pic at \RightBC{\BotY}{\ShiftRight}            {trap={orange!30}{H}};
    \pic at \RightBC{\BotY}{\ShiftRight+\BotStep}   {trap={orange!30}{I}};
    \pic at \RightBC{\BotY}{\ShiftRight+2*\BotStep} {trap={orange!30}{J}};
    \node[below=4pt] (HqLabel) at ($ (H_bc)!0.5!(J_bc) $) {$\mathbf{H}_q$};

    \node[below=1.4cm of HqLabel] (LangInstrAnchor) {};
    \node[below=1.4cm of HqLabel, xshift=-1.5cm] (LangInstr) {Language Instruction $\mathbf{X}_q$};

    \draw[->, thin] (LangInstrAnchor.north) -- (HqLabel.south);

    \tikzset{ga/.style={->, very thin, draw=gray!70}}
    \draw[ga] (E_tc) -- (A_bc);
    \draw[ga] (F_tc) -- (A_bc);
    \draw[ga] (F_tc) -- (B_bc);
    \draw[ga] (G_tc) -- (A_bc);
    \draw[ga] (G_tc) -- (B_bc);
    \draw[ga] (G_tc) -- (C_bc);
    \foreach \s in {H,I,J} {
        \draw[ga] (\s_tc) -- (A_bc);
        \draw[ga] (\s_tc) -- (B_bc);
        \draw[ga] (\s_tc) -- (C_bc);
      }


    \draw[->, thin](projw.east) -- (HvLabel.west);
  \end{tikzpicture}
  \caption{Our experimental approach with LLaVA Framework \cite{liu2023llava} that compares the performance of different image encoders and 2D-RoPE variants. }
  \label{fig:llavaarch}
\end{figure}


\subsection{Training}
Models are trained using the standard two-stage LLaVA recipe: projection pretraining followed by full instruction tuning, using the original LLaVA datasets. Images are resized to $256 \times 256$. Pretraining updates only the projection layer, while instruction tuning updates all parameters. Full training and optimization details are provided in the appendix.

\section{Results}
\label{sec:results}

We evaluate frontier multimodal models and controlled LLaVA-1.5 (7B) variants across spatial reasoning benchmarks. All encoder-swapped models and their 2D-RoPE counterparts are trained on the same 7B LLaVA backbone for controlled comparison. Frontier models between 2B and 8B parameters include LLaVA-NeXT \citep{liu2024llavanext}, LLaVA-OneVision-qwen2-7B-ov-hf \citep{li2024llavaonevisioneasyvisualtask}, Qwen2.5-VL-8B \citep{bai2025qwen25vltechnicalreport}, SmolVLM2-2.2B-Instruct \citep{marafioti2025smolvlm}, Gemma3-4b-it \citep{gemmateam2025gemma3technicalreport}, PaliGemma2-3b-mix-448 \citep{paliGemma2024}, and Molmo-7B-D-0924 \citep{deitke2024molmo}.

\begin{table*}[t]
    \centering
    \caption{Comparison of frontier models and LLaVA variants across spatial understanding benchmarks. Values \underline{underlined} indicate the best-performing frontier model; values in \textbf{bold} indicate the best-performing LLaVA variant.}
    {\setlength{\tabcolsep}{4pt} 
    \begin{tabular}{lccccccc}
        \toprule
        Models & MMVP & \makecell{CV-Bench \\ 2D Overall} & TallyQA & \makecell{GQA \\ Overall} & VSR & \makecell{Top-\\ViewRS} & \makecell{Count-\\BenchQA} \\
        \toprule
        LLaVA-NeXT & 0.667 & 0.606 & 0.733 & \underline{63.786} & 63.994 & 0.409 & 0.515 \\
        \nobrk{LLaVA-OneVision} & 0.767 & 0.730 & 0.797 & 62.140 & 77.741 & 0.414 & 0.823 \\
        Qwen2.5-VL & \underline{0.770} & \underline{0.754} & 0.800 & 60.391 & \underline{89.116} & \underline{0.456} & \underline{0.891} \\
        SmolVLM2 & 0.687 & 0.577 & 0.729 & 50.574 & 71.277 & 0.416 & 0.692 \\
        \nobrk{Gemma3-4b-it} & 0.708 & 0.659 & 0.525 & 31.277 & 55.074 & 0.334 & 0.713 \\
        PaliGemma & 0.667 & 0.624 & 0.794 & 62.570 & 65.139 & 0.322 & 0.674 \\
        Molmo & 0.753 & 0.728 & \underline{0.808} & 55.295 & 76.432 & 0.323 & 0.858 \\
        LLaVA v1.5 & 0.577 & 0.490 & 0.707 & 33.225 & 55.810 & 0.384 & 0.468 \\
        \midrule
        \nobrk{LLaVA-2D-RoPE} & 0.513 & 0.443 & 0.654 & 34.433 & 57.201 & 0.283 & 0.290 \\
        LLaVA-SigLIP & 0.433 & 0.412 & 0.672 & 25.648 & 54.910 & 0.349 & 0.581 \\
        \nobrk{LLaVA-SigLIP-2D-RoPE} & 0.507 & 0.425 & 0.616 & \textbf{38.448} & 57.692 & 0.295 & 0.483 \\
        LLaVA-SigLIP2 & 0.427 & 0.442 & 0.684 & 23.970 & 52.701 & \textbf{0.371} & 0.532 \\
        \nobrk{LLaVA-SigLIP2-2D-RoPE} & 0.480 & 0.415 & 0.646 & 34.560 & 56.465 & 0.330 & 0.402 \\
        LLaVA-AIMv2 & 0.513 & \textbf{0.466} & \textbf{0.710} & 32.541 & 56.219 & 0.339 & \textbf{0.739} \\
        \nobrk{LLaVA-AIMv2-2D-RoPE} & \textbf{0.560} & 0.432 & 0.690 & 32.342 & \textbf{60.311} & 0.338 & 0.719 \\
        \bottomrule
    \end{tabular}
    }
    \label{tab:vlm-spatial--benchmarks}
\end{table*}

\subsection{Frontier models still fail on spatial reasoning}
Table~\ref{tab:vlm-spatial--benchmarks} shows that Qwen2.5-VL is the strongest frontier model overall, leading on CV-Bench 2D Overall, MMVP, VSR, TopViewRS, and CountBenchQA, while LLaVA-NeXT leads frontier models on GQA Overall and Molmo leads on TallyQA. Despite these gains, spatial performance remains uneven across tasks and models, indicating that spatial grounding is not consistently captured by general-purpose training and scaling.

\subsection{Encoder choice dominates spatial performance in controlled LLaVA variants}
Within the controlled LLaVA setting, we observe that the choice of image encoder strongly shapes spatial behavior. LLaVA-AIMv2 yields the most consistent improvements over the CLIP-based LLaVA baseline, achieving the best LLaVA scores on CV-Bench 2D Overall, TallyQA, and CountBenchQA. LLaVA-AIMv2-2D-RoPE further improves MMVP and achieves the best LLaVA score on VSR. In contrast, SigLIP- and SigLIP2-based variants show more fragmented gains, with LLaVA-SigLIP-2D-RoPE performing best on GQA Overall and LLaVA-SigLIP2 performing best on TopViewRS.

Overall, these results suggest that spatial failures in VLMs are strongly coupled to the visual backbone: encoders optimized primarily for global image--text alignment do not reliably support spatial reasoning, while encoders trained with denser or generative objectives can improve spatial outcomes within the same multimodal framework.

\subsection{Qualitative localization reveals missing spatial grounding}

\begin{figure}[t]
  \centering
  \begin{subfigure}[t]{0.45\linewidth}
    \centering
    \begin{subfigure}{0.45\linewidth}
      \centering
      \fbox{\includegraphics[width=\linewidth,height=\linewidth]{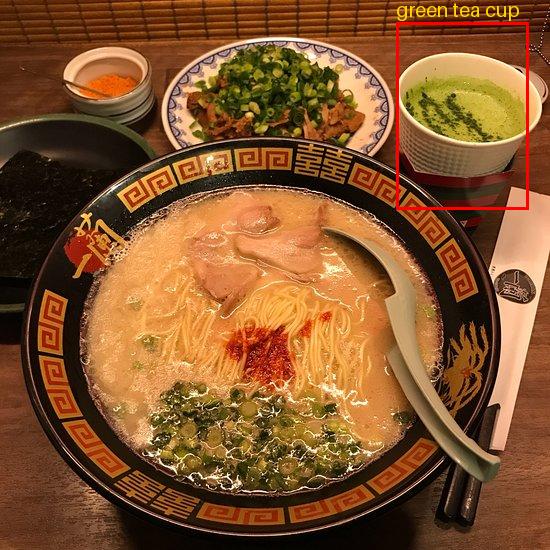}}
      \caption{SigLIP2}
      \label{fig:suba}
    \end{subfigure}
    \hfill
    \begin{subfigure}{0.45\linewidth}
      \centering
      \fbox{\includegraphics[width=\linewidth,height=\linewidth]{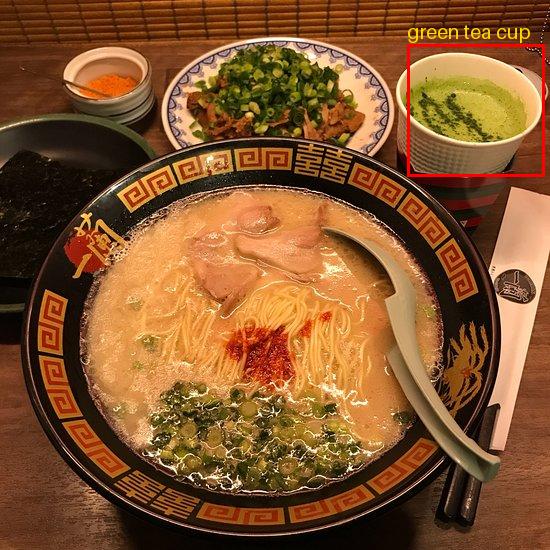}}
      \caption{AIMv2}
      \label{fig:subb}
    \end{subfigure}
    \caption{Prompt: \textit{Locate the cup that contains green liquid. Provide the bounding boxes.}}
    \label{fig:group1}
  \end{subfigure}
  \hfill
  \begin{subfigure}[t]{0.45\linewidth}
    \centering
    \begin{subfigure}{0.45\linewidth}
      \centering
      \fbox{\includegraphics[width=\linewidth,height=\linewidth]{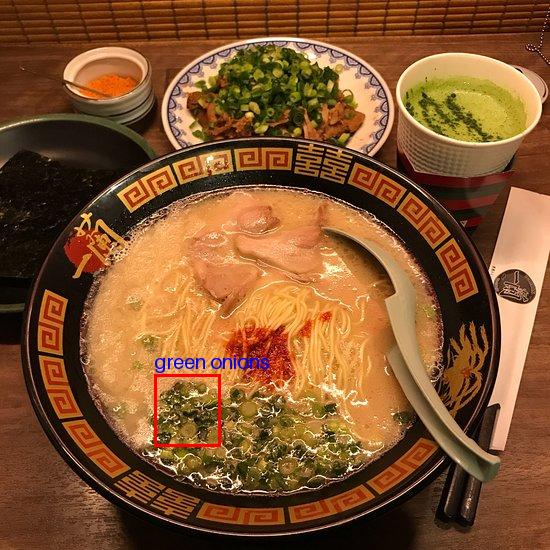}}
      \caption{SigLIP2}
      \label{fig:subc}
    \end{subfigure}
    \hfill
    \begin{subfigure}{0.45\linewidth}
      \centering
      \fbox{\includegraphics[width=\linewidth,height=\linewidth]{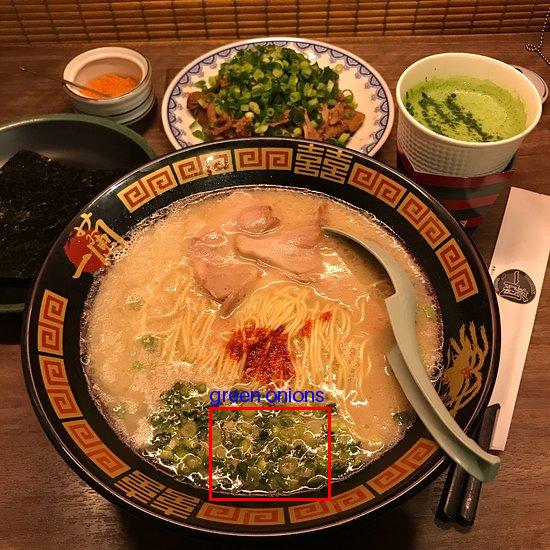}}
      \caption{AIMv2}
      \label{fig:subd}
    \end{subfigure}
    \caption{Prompt: \textit{Locate the pieces of green onions in the ramen bowl. Provide the bounding box coordinates given the image size is 550$\times$550.}}
    \label{fig:group2}
  \end{subfigure}

  \caption{Object localization in LLaVA-SigLIP2 vs. LLaVA-AIMv2. AIMv2 yields tighter and more spatially aligned localizations, while SigLIP2 often produces imprecise or misaligned boxes.}
  \label{fig:bounding_box_siglip2_aimv2}
\end{figure}

Quantitative results suggest that encoder choice strongly affects spatial behavior. To further examine this, we qualitatively compare LLaVA-SigLIP2 and LLaVA-AIMv2 on object localization prompts. Figure~\ref{fig:bounding_box_siglip2_aimv2} shows representative examples where models are asked to localize visually grounded regions and output bounding boxes.

Across examples, LLaVA-AIMv2 produces tighter and more accurate localizations, while LLaVA-SigLIP2 frequently outputs imprecise or spatially misaligned boxes. These failures are consistent with the quantitative trends on CV-Bench and VSR, and suggest that encoders trained with dense or generative supervision better preserve spatial detail needed for grounded perception. In contrast, encoders optimized primarily for global alignment struggle to support region-level reasoning, even when paired with the same language model and training pipeline.

\subsection{2D positional structure helps, but does not resolve spatial grounding}
We also evaluate the impact of 2D-RoPE across encoder variants. Improvements are mixed: 2D-RoPE helps in some settings (e.g., AIMv2 on MMVP and VSR; SigLIP on GQA Overall), but degrades others (e.g., CV-Bench 2D Overall for AIMv2, and several tasks for the CLIP baseline). These results indicate that preserving 2D positional structure alone is not sufficient; spatial grounding depends jointly on the visual features learned by the encoder and the positional structure used during multimodal fusion.

\begin{figure*}[t]
  \centering
  \begin{minipage}[t]{0.2\textwidth}
    \centering
    \captionof{figure}{Example image from LLaVA-Bench (In-the-Wild) \citep{liu2023llava}.}
    \label{fig:asianfood_vqa}
    \vspace{0.4\baselineskip}
    \fbox{\includegraphics[width=\linewidth]{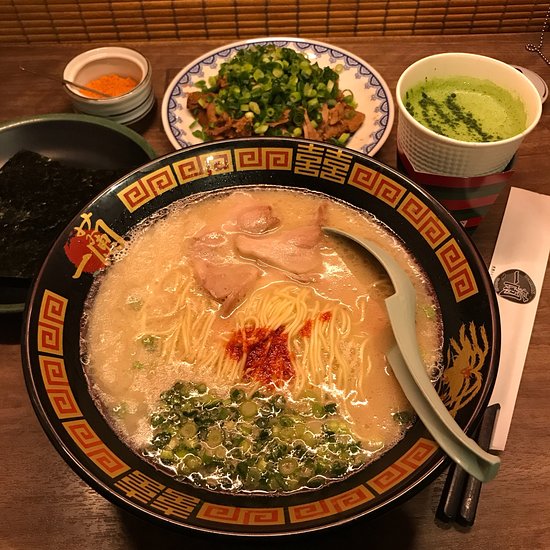}}
  \end{minipage}
  \hfill
  \begin{minipage}[t]{0.77\textwidth}
    \centering
    \captionof{table}{Model outputs for the prompt \textit{Are the chopsticks to the left or right of the bowl?} on image shown in Figure~\ref{fig:asianfood_vqa}.}
    \label{tab:asianfood_vqa_results}
    \begin{tabular}{l l}
        \toprule
        \textbf{Model} & \textbf{Output} \\
        \midrule
        LLaVA-v1.5 & The chopsticks are to the right of the bowl. \\
        LLaVA-2D-RoPE & The chopsticks are to the right of the bowl. \\
        LLaVA-SigLIP & The chopsticks are to the right of the bowl. \\
        LLaVA-SigLIP-2D-RoPE & The chopsticks are to the right of the bowl. \\
        LLaVA-SigLIP2 & The chopsticks are to the right of the bowl. \\
        LLaVA-SigLIP2-2D-RoPE & Right \\
        LLaVA-AIMv2 & The chopsticks are to the right of the bowl. \\
        LLaVA-AIMv2-2D-RoPE & The chopsticks are to the right of the bowl. \\
        Qwen2.5-VL & The chopsticks are to the right of the bowl. \\
        \textbf{Gemma3-4b-it} & \textbf{The chopsticks are to the left of the bowl.} \\
        \bottomrule
    \end{tabular}
  \end{minipage}
\end{figure*}

\section{Conclusion}
\label{sec:conclusion}

Despite rapid progress in VLMs, spatial reasoning remains fragile and inconsistent. Even frontier models vary widely across spatial benchmarks, and strong performance on general multimodal tasks does not reliably translate into spatial grounding. In a controlled LLaVA setting, we show that architectural choices—particularly the image encoder objective and multimodal positional structure shape spatial behavior. Encoders trained with denser or generative supervision improve spatial performance, while 2D positional structure alone is insufficient. Overall, our results suggest that spatial reasoning is an under-addressed design dimension in modern VLMs. We hope this diagnostic study encourages treating spatial representation as a first-class concern in VLM design and evaluation.

\section{Future Work}
\label{sec:futurework}

Our study focused on static, 2D images, benchmarks, and encoder variants within the LLaVA framework. This work can extend to 3D spatial reasoning in conjunction with the dynamic environment. Another potential extension can be on SigLIP2 with NaFlex. The flexible resolution image preprocessing of NaFlex mitigates the information loss observed in fixed-resolution encoders. In terms of visual backbones, incorporating DINOv2 in LLaVA is left out for future work. Future work will analyze why 2D-RoPE sometimes degrades performance by adding attention-based diagnostics and testing for potential conflicts with the LLM’s positional encoding. We will introduce spatial probing of encoder and fused representations to identify what spatial information is present and where it is lost in the pipeline. Another area is to explicitly study image resolution as a confounder by running controlled experiments beyond the current 256×256 setting and reporting its independent effect on spatial reasoning.

\section{Acknowledgment}
\label{sec:ack}
We thank EleutherAI for their generous GPU support. We also thank our mentor, Jekaterina Novikova, for asking critical questions that strengthened this project.



\clearpage
\bibliography{iclr2026_conference}
\bibliographystyle{iclr2026_conference}


\end{document}